\documentclass[10pt,twocolumn,letterpaper]{article}

\usepackage{cvpr}
\usepackage{times}
\usepackage{epsfig}
\usepackage{graphicx}
\usepackage{amsmath}
\usepackage{amssymb}
\usepackage{amsmath}
\usepackage{amssymb}
\usepackage{algorithm}
\usepackage{algorithmic}
\usepackage{comment}
\usepackage{booktabs}
\usepackage{subcaption}

\usepackage[pagebackref=true,breaklinks=true,letterpaper=true,colorlinks,bookmarks=false]{hyperref}

\cvprfinalcopy 

\def\cvprPaperID{3097} 

\ifcvprfinal\fi
\begin{document}

\title{Cross-Channel Intragroup Sparsity Neural Network}

\author{
Zilin Yu\thanks{Equal Contribution. Corresponding author: X. Wu}\\
Hangzhou Dianzi University\\
kavioyu@gmail.com\\
\and
Chao Wang$^*$\\
Peiking University\\
wangchao16@pku.edu.cn\\
\and
Xin Wang\\
Cerebras Systems\\
poincare.disk@gmail.com\\
\and
Qing Wu\\
Hangzhou Dianzi University\\
wuq@hdu.edu.cn\\
\and
Yong Zhao\\
Peiking University\\
yongzhao@pkusz.edu.cn\\
\and
Xundong Wu\\
Hangzhou Dianzi University\\
wuxundong@gmail.com}

\maketitle

\begin{abstract}
Modern deep neural networks rely on overparameterization to achieve state-of-the-art generalization. But overparameterized models are computationally expensive. Network pruning is often employed to obtain less demanding models for deployment. Fine-grained pruning removes individual weights in parameter tensors and can achieve a high model compression ratio with little accuracy degradation. However, it introduces irregularity into the computing dataflow and often does not yield improved model inference efficiency in practice. Coarse-grained model pruning, while realizing satisfactory inference speedup through removal of network weights in groups, \textit{e.g.} an entire filter, often lead to significant accuracy degradation. This work introduces the cross-channel intragroup (CCI) sparsity structure, which can prevent the inference inefficiency of fine-grained pruning while maintaining outstanding model performance. We then present a novel training algorithm designed to perform well under the constraint imposed by the CCI-Sparsity. Through a series of comparative experiments we show that our proposed CCI-Sparsity structure and the corresponding pruning algorithm outperform prior art in inference efficiency by a substantial margin given suited hardware acceleration in the future.
\end{abstract}


\section{Introduction} \label{intro}
State-of-the-art performance of deep neural networks in many computer vision tasks has set off the trend of real-world deployment of these models. Though it is desirable to use networks of the best performance, superior generalization often requires high computing complexity due to large network widths and depths. The resulting high cost and latency are often prohibitive for resource-limited platforms, such as mobile devices.

To improve computational efficiency, various methods have been proposed to produce lightweight network architectures while maintaining satisfactory model performance \cite{han2015learning,howard2017mobilenets}. A widely used technique, \emph{network pruning} removes unimportant weights to yield sparse models of lower computational complexity. Pruning can be either \emph{fine-grained}, \textit{i.e.} individual weights are independently targeted for removal \cite{han2015learning}, or \emph{structured} (coarse-grained), \textit{i.e.} weights are removed in groups, such as entire channels or blocks of weights inside a filter \cite{mao2017exploring}. Fine-grained pruning typically yields models with higher parameter efficiency; however, it often does not improve computational efficiency at inference time due to the data accessing irregularity resulting from haphazard sparsity patterns. Structured pruning, however, can be constructed to realize improved inference efficiency, but it often leads to further performance degradation \cite{mao2017exploring}.

\begin{figure}[htb]

\centering
\begin{subfigure}[t]{0.44\columnwidth}
\centering
\includegraphics[width=\textwidth]{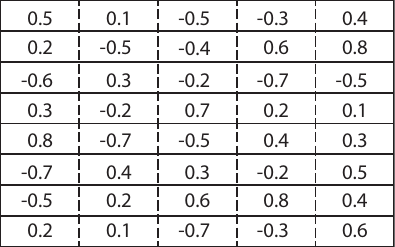}
\caption{}\label{fig:dense_weight}
\end{subfigure}
\hfill
\begin{subfigure}[t]{0.44\columnwidth}
\centering
\includegraphics[width=\textwidth]{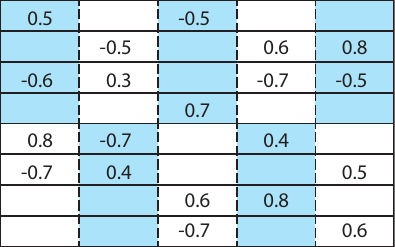}
\caption{}\label{fig:sparse_weight}
\end{subfigure}
\caption{(a) An illustration of a weight matrix prior to pruning. An output channel corresponds to one of the 8 rows contains 5 weights each, \textit{i.e.} 5 input features transformed into 8 output features. (b) The weight matrix undergone pruning with CCI-Sparsity constraint. In this case, four contiguous weights (illustrated as a block of the same color) from four neighboring channels form one weight group, \textit{i.e.} group size is 4, and there are exactly 2 nonzero weights per group, resulting in a sparsity level of 50\%.}
\label{fig:weight}
\end{figure}

To push beyond the frontier set by this tradeoff and achieve both high accuracy and computational efficiency, this study introduces \emph{cross-channel intragroup} (CCI) sparsity (Fig. \ref{fig:weight}), a sparseness pattern designed to avoid the irregularity in inbound data flow, \textit{viz.} in reading of the input and weight tensors from memory, of a fine-grained sparse parameter tensor. In contrast to unconstrained fine-grained sparse network, weights in each network layer with CCI-Sparsity structure are subdivided into small groups such that active weight pruning yields a fixed number of nonzero weights for each group. The weight groups are arranged to be contiguous along network output channels.

Our experimental results suggest that the weight group size can be made small (8 and 16) without significantly degrading the model's generalization performance, substantially outperforming structured pruning. At the same time, the proposed CCI-Sparsity structure eliminates the data inflow irregularity associated with unconstrained fine-grained sparse networks, which, with proper hardware acceleration, enables speedup of network inference that scales linearly with the model sparsity level.

\paragraph{Contributions:} 
\begin{enumerate}
    \item   We propose the CCI-Sparsity structure, that enables efficient sparse neural network inference while maintaining the performance advantage of the fine-grained sparsity over structured pruning.

    \item  We present theoretical analysis on how weight group size affects sparsification, and on how sparsity level affects performance of pruned networks with CCI-Sparsity or Balanced-Sparsity. 

    \item  We propose a solution that can overcome the difficulty of training networks with CCI-Sparsity (or Balanced-Sparsity \cite{yao2019balanced}) structure with small weight group sizes. Our approach outperforms the iterative model pruning approach employed in Yao \etal \cite{yao2019balanced}. Our method produces models substantially outperform numerous state-of-the-art lightweight architectures and pruned models with structured sparsity.

    \item  We analyze inference I/O complexity of CCI-sparsity and Balanced-Sparsity structures, and demonstrate the advantage of CCI-Sparsity under mild assumptions.

\end{enumerate}

The rest of this article is organized as follows. 
Sec. \ref{related} summarizes relevant literature. 
Sec. \ref{Section:CI} defines the CCI-Sparsity structure and explains the resulting efficiency for inference. 
Sec. \ref{Section:constraint} gives a theoretical analysis on how CCI-Sparsity constraints affect model performance. 
Sec. \ref{Section:drop} describes the pruning algorithm designed to optimize performance under the CCI-Sparse constraint. This section also demonstrates the advantage of our pruning algorithm over the method employed in Yao \etal \cite{yao2019balanced}. 
Sec. \ref{Section:coarse_comp} demonstrates the advantage of CCI-Sparsity over structured sparsity and other lightweight architectures. 
Sec. \ref{Section:conclusion} concludes the article.
\section{Related Work} \label{related}


\subsection{Lightweight CNN architectures}

Convolutional neural networks (CNNs) are inherently sparse in that each convolutional filter only receives inputs from a small spatial neighborhood, engendering an advantage over fully connected networks, for image processing, in both efficiency and performance. In some widely adopted CNN models, such as AlexNet and VGG, convolutional filters receive inputs of a rather high dimensionality, making convolutional parameters contain significant redundancy.

Tensor decomposition curtails redundancy through approximating network filters with low-rank tensors, thereby reducing model complexity. Recent lightweight architectures such as MobileNet \cite{howard2017mobilenets}, MobileNetV2 \cite{sandler2018mobilenetv2}, decompose convolutional layers into filters on different spatial dimensions. Specifically, they employ depth-wise separable convolutions in place of regular convolutional filters, reducing the computation complexity. 
Group convolution \cite{gao2018channelnets,huang2018condensenet,ma2018shufflenet,zhang2017interleaved} goes one step further by separating convolution operations into groups, in effect removing the weight connections between filters that are not in the same group. The proposed CCI-Sparsity structure in this work has the advantage over group convolution in that it reduces input dimensionality while not suffering from the strong constraint of separating input into groups.

\subsection{Sparse CNNs}
Reducing network redundancy through pruning (or sparsification) of a large network is an alternative for constructing compact networks. Unstructured pruning removes individual network weights that meet certain criteria \cite{han2015learning} and generates networks of fine-grained sparsity. It is highly effective in removing redundant network weights, thus can greatly reduce model sizes. However, due to their unstructured nature, fine-grained sparse models can often be inefficient at inference time.

Structured pruning removes network connections under constraints that allow efficient model inference. Those constraints require removal of contiguous blocks of connections instead of individual weights. The unit of pruning block can be a filter, a channel or a sub-block of weights inside filters \cite{li2016pruning,mao2017exploring,narang2017block,wen2016learning}. However, those structured pruning procedures often lead to substantial accuracy degradation \cite{mao2017exploring,yao2019balanced}. 

\subsection{Intragroup sparsity}
Intragroup sparsity \cite{zhou2010exclusive} is originally proposed as a form of model regularization, where the $\ell_{1,2}$-norm is used to induce sparsity at the intragroup level for model feature selection. In Wu \etal \cite{wu2018improved}, the authors adopt intragroup sparsity structure to overcome the irregularity associated with their dendritic neural networks. In their study, the weight group is constructed across dendritic subkernels with \textit{fixed} connection maps. Their proposed intragroup sparsity structure remains mostly conceptual without detailed analysis. In this study, weight groups are constructed across network channels with \textit{learned} connection maps. 

\subsection{Balanced-Sparsity}
Another related approach to the same end is Balanced-Sparsity \cite{yao2019balanced}. In contrast to CCI-Sparsity, the weight groups in Balanced-Sparsity are formed by contiguous weights from inside the same, instead of across different, output channels (weights from the same rows instead of columns in Fig. \ref{fig:weight}). 
The group sizes used in their study are much larger than in this work. Our work also proposes a training algorithm that significantly outperforms the iterative pruning technique employed by Yao \etal, and generates more efficient models with small block sizes. We discuss this further in Sec. \ref{Section:CCI_BBS}.  
Moreover, here we show that CCI-Sparsity, under mild assumptions, is more conducive for hardware acceleration.

\section{CCI-Sparsity} \label{Section:CI}
Pruned networks with unstructured, fine-grained sparsity patterns often lead to model inference inefficiency on popular hardware accelerators \cite{cao2019efficient,han2016eie,mao2017exploring,parashar2017scnn,zhang2016cambricon,zhou2018cambricon}.

\subsection{CCI-Sparsity structure}
To address the computing inefficiency associated with fine-grained sparse networks, we propose CCI-Sparsity. CCI-Sparsity is compatible with both convolutional and fully connected networks. For clarity, here we use a fully connected neural network layer as an example to explain the concept. Consider a fully connected layer, we represent its input by column vector $X \in\mathbb{R}^N$, its weights by a matrix $W \in \mathbb{R}^{M\times N}$ and its output column vector $H\in\mathbb{R}^M$, s.t. $H=W\times{X}$ is satisfied. For example, for the weight matrix $W$ illustrated in Fig. \ref{fig:dense_weight}, we have $M=8, N=5$, and each row of $W$ contains weights for one output channel in $H$. For the CCI-Sparsity configuration, the weights are partitioned into groups. For each weight group, a fixed number of weights are set to zero during sparsification. Fig. \ref{fig:sparse_weight} illustrates the corresponding weight matrix, denoted by $\hat{W}$, that is obtained after pruning with a weight group size of $G=4$ and $s=2$ nonzero weights per group. The weights for each group are from 4 channels (rows in $W$). As such, we can compress every $G=4$ rows in $\hat{W}$ into exactly $s=2$ rows of compressed weights. Each new row in $\hat{W}$ will have the same number of columns as the original weight matrix.

To identify the row in which a weight in $\hat W$ is located within $W$, we assign a row index to each weight in the compressed matrix (this index also specifies to which row in $H$ we add the multiplication result). For this example of $G=4$, we require a 2-bit index for each weight. With $G=4$ and $s=2$, we require 4 index bits for each weight group--even though the theoretical number of index bits is slightly smaller, it is practical to use the plain index to avoid extra decoding overhead. For an arbitrary group size of $G$, a plain weight index will consist of $log_2G$ bits. Small group sizes incur lower overhead in the extra storage and the I/O requirements associated with weight indices, while at the same time place stronger constraints on the model, yielding lower accuracy (see Sec. \ref{Section:drop}).

\subsection{Efficient inference through CCI-Sparsity} \label{anal}
To demonstrate that CCI-Sparsity can eliminate irregular data access so as to enable efficient neural network inference, let us consider a standard matrix-matrix multiplication case, namely, $H=W\times X$, with weight matrix $W$, input $X$, and multiplication result $H$. Matrix multiplication of this form is often called general matrix-matrix multiplication (GEMM), which is the core computation routine that is heavily used in neural network inference for both convolutional and fully connected network architectures \cite{chetlur2014cudnn}. First, we consider $W, X, H \in\mathbb{R}^{N\times N}$ in their dense forms, as shown in Fig. \ref{fig:CCI_T}. We present a naive form of the GEMM as in the following algorithm. The number of computational steps of the naive procedure for this case is $N^3$. 

\begin{algorithm}
  \caption{Dense matrix-matrix multiplication. Accessing to $H, W,\, and\, X$ is sequential, with $H,W$ stored in row major, $X$ in column major.}
  \label{alg0}
  \begin{algorithmic}
  \REQUIRE ${W}, {X} \in \mathbb{R}^{N\times N}$, ${H}=0$ .
  \ENSURE ${H}= {W}\times {X}$
  \STATE \textbf{for}\ {$i \gets 0$ to $N$ by $1$}
    \STATE \quad \textbf{for}\ {$j \gets 0$ to $N$ by $1$}
        \STATE \quad \quad \textbf{for}\ {$k \gets 0$ to $N$ by $1$}
                    \STATE \quad \quad \quad ${H}(i,j) \gets {H}(i,j)+{W}(i,k)\times{{X}(k,j)}$
  \end{algorithmic}
\end{algorithm}

Next, let us compress the weight matrix $W$ into CCI-Sparsity format $\hat W$. In this case, if we assume a group size of $G=4$ with $s=1$, we reduce the size of the weight matrix by a factor of $G/s=4$ times. Each element in $\hat W$ now has two components: a weight value and an index. We denote them as $\hat W(i,k)[V]$ and $\hat W(i,k)[I]$, respectively. We compute them via Alg. \ref{alg1} (for clarity, we assume $s=1$ here). With CCI-Sparsity, we can reduce the number of computational steps by a factor of $G$ to $N^3/G$. At the same time, with CCI-Sparsity, the reading dataflows of $\hat W$ and $X$ remains continuous, as in the dense case. The dataflow of matrix $H$ is also continuous at the group block level. Irregularity only remains inside each step when the $Idx$ associated with each weight is used to access the corresponding element inside the output matrix group. This can be overcome through loading a whole group of elements of $H$ onto on-chip registers, thereby circumventing the latency and inefficiency that are caused by irregular off-chip memory access \cite{cao2019efficient}. 

\begin{algorithm}
  \caption{CCI-Sparse matrix-matrix multiplication. Accessing to $\hat W, X$ is sequential, with $H, \hat W$ stored in row major, $X$ in column major.}
  \label{alg1}
  \begin{algorithmic}
  \REQUIRE ${\hat W}, {X} \in \mathbb{R}^{N\times N}$, ${H}=0$ .
  \ENSURE ${H}= {\hat W}\times {X}$
  \STATE \textbf{for}\ {$i \gets 0$ to $N/G$ by $1$}
    \STATE \quad \textbf{for}\ {$j \gets 0$ to $N$ by $1$}
        \STATE \quad \quad \textbf{for}\ {$k \gets 0$ to $N$ by $1$}
                    \STATE \quad \quad \quad $(Idx, Val) \gets ({\hat W}(i,k)[I], {\hat W}(i,k)[V])$
                    \STATE \quad \quad \quad ${H}(G\times i+Idx,j) \gets {H}(G\times i+Idx,j)+Val\times{{X}(k,j)}$
  \end{algorithmic}
\end{algorithm}


Thus, CCI-Sparsity can improve the dataflow of regular fine-grained sparse networks. Furthermore, thanks to the introduction of the weight group structure, CCI-Sparsity enables us to split execution along the boundaries of weight groups, enabling parallel processing and matrix-tilling-based data reuse for improving inference efficiency \cite{cao2019efficient,yao2019balanced,zhou2018cambricon}.

\subsection{CCI-Sparsity vs. Bank-Balanced Sparsity}
\label{Section:CCI_BBS}

\begin{figure}[htb]
\centering
\begin{subfigure}[t]{0.49\columnwidth}
\centering
   \includegraphics[width=\textwidth]{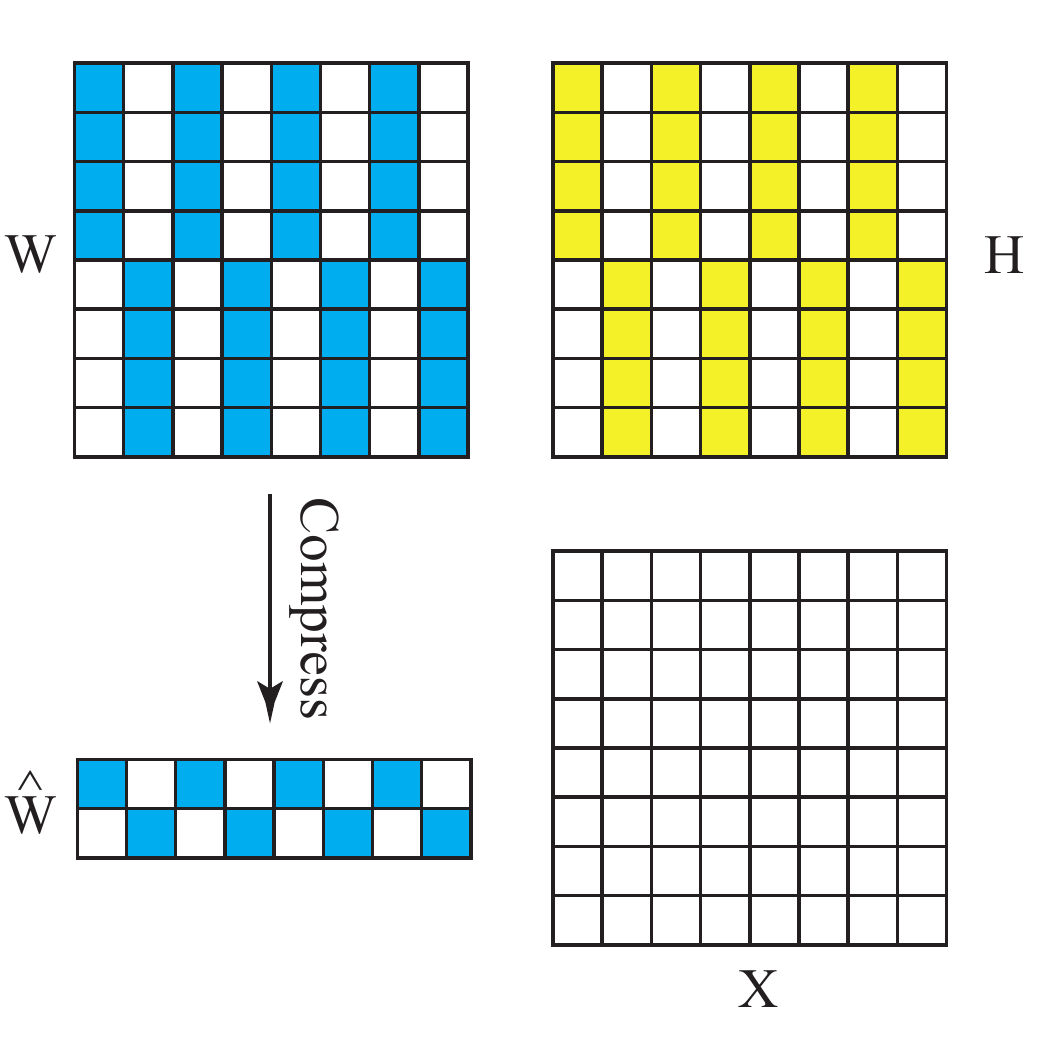}

   \caption{}
\label{fig:CCI_T}
\end{subfigure}
\hfill
\begin{subfigure}[t]{0.49\columnwidth}
\centering
   \includegraphics[width=\textwidth]{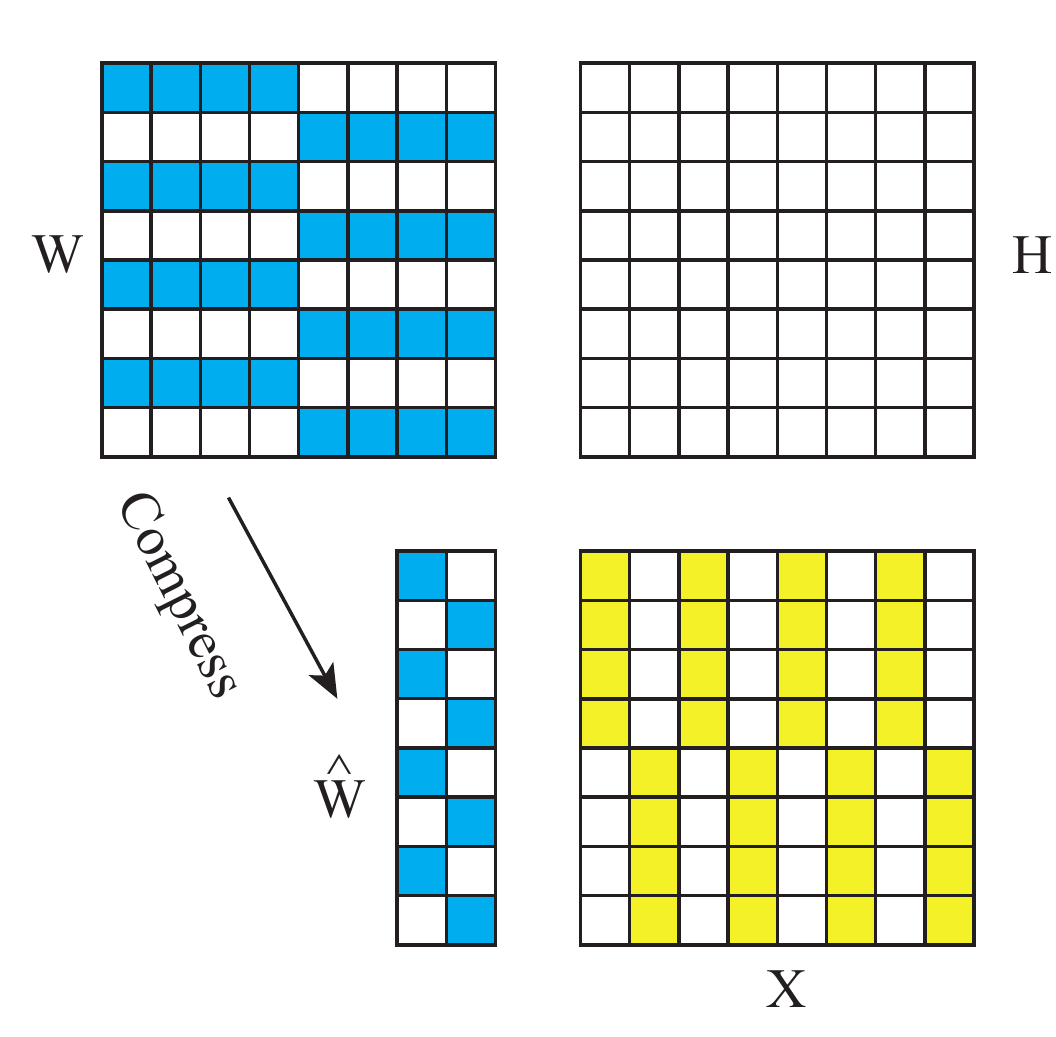}

   \caption{}
\label{fig:BBS_T}
\end{subfigure}
\begin{subfigure}[t]{0.49\columnwidth}
\centering
   \includegraphics[width=\textwidth]{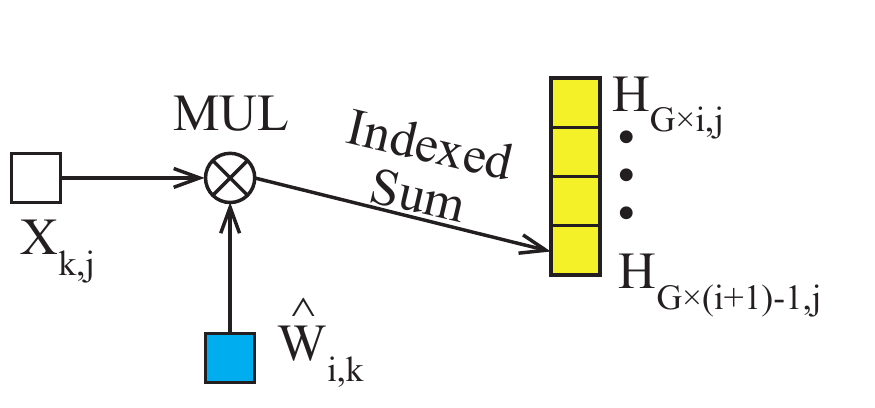}

   \caption{}
\label{fig:CCI_B}
\end{subfigure}
\hfill
\begin{subfigure}[t]{0.49\columnwidth}
\centering
   \includegraphics[width=\textwidth]{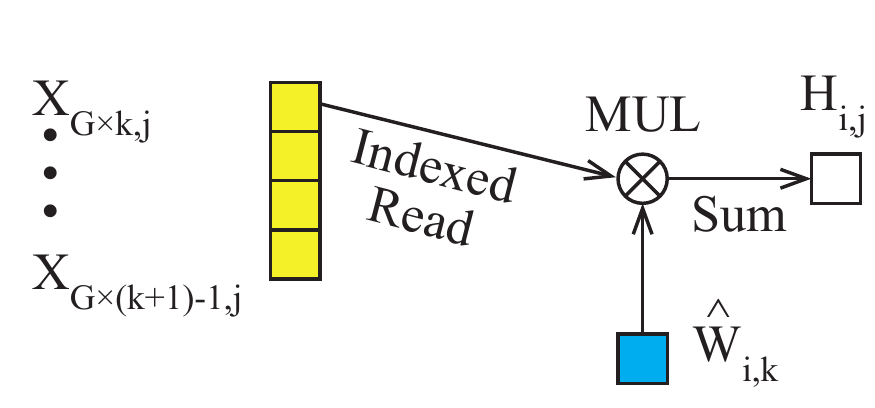}

   \caption{}
\label{fig:BBS_B}
\end{subfigure}
\caption{Procedural difference in GEMM between CCI-Sparsity and Balanced-Sparsity.  (a) Matrix multiplication with CCI-Sparsity. (b) Matrix multiplication with Balanced-Sparsity. (c) Illustration of a basic processing element for CCI-Sparsity. (d) Illustration of a basic processing element for Balanced-Sparsity.}
\label{fig:Compare_all}
\end{figure}

In \cite{cao2019efficient,yao2019balanced}, the authors propose and demonstrate the performance of the Balanced-Sparsity structure. CCI-Sparsity is similar to Balanced-Sparsity in that both introduce uniformly sized weight groups in network parameters, and both associate an index with each compressed weight. The central difference between the two structures is illustrated in Fig. \ref{fig:Compare_all}. In CCI-Sparsity, weight groups are formed between weights that correspond to different output channels ($G=4$ contiguous weights of the same color, as in Fig. \ref{fig:CCI_T}). For example, $W_{0,0},W_{1,0},W_{2,0},\, \text{and} \,W_{3,0}$ form a weight group, and they are compressed into a single weight in $\hat W$ as $\hat W_{0,0}$. Assume $W_{2,0}$ is the weight that survives the pruning process; that is, a weight index of 2 is associated with $\hat{W}_{0,0}$. Then, the weight value of $\hat W_{0,0}$ is multiplied with the $X_{0,i}$, and accumulated to the corresponding $H_{2,i}$. This computational procedure is illustrated in Fig. \ref{fig:CCI_B}.

For Balanced-Sparsity \cite{yao2019balanced}, weight groups are formed by weights that correspond to different inputs ($G=4$ contiguous weights of the same color, as illustrated in Fig. \ref{fig:BBS_T}). Thus, the index that is associated with the compressed weights is used to allocate the corresponding input elements inside a group, as illustrated in Fig. \ref{fig:BBS_B}.

Numerous recent studies on specialized neural network accelerators show that data I/O, namely reading/writing data from/to off-chip memory, dominates the total energy budget \cite{zhou2018cambricon}. The dataflow designs differ drastically among accelerators so as to optimize system efficiency for specific use cases \cite{deng2020model}. 
While a thorough analysis of the dataflow optimization for either CCI or Balanced-Sparsity on all accelerator designs is beyond the scope of this study, we analyze a case with mild assumptions demonstrating a clear advantage of CCI-Sparsity over Balanced-Sparsity.

In this analysis, we assume that each processing element (PE) contains the minimum number of registers that are necessary for the implementation of either CCI-Sparsity or Balanced-Sparsity. No matrix tiling or other data parallelism is considered here. For the CCI-Sparsity case, we use the output stationary dataflow; that is, for each stride, we set up G registers for the output summation results of a group (contiguous blocks of the same color, as in matrix $H$ of Fig. \ref{fig:CCI_T}). Then, we keep those $G$ registers stationary to maximum their reuse. Inside a stride, for each calculation step, we read one element from $\hat{W}$ and one element from $X$. There are $N$ steps inside each stride. At the end of a stride, we write $G$ elements of $H$ into memory. The total number of groups inside $H$ is $N^2/G$. Thus, the memory I/O requirement for the multiplication is $(2N+G)\times N^2/G=2N^3/G+N^2$. In the case of Balanced-Sparsity, as illustrated in Fig. \ref{fig:BBS_B}, for each stride, we first read $G$ elements from $X$ and store them in registers for maximum reuse. Then, for each computational step, we also have to read one element from $\hat W$, whereas for $H$, we not only need to read the element but also to write it, because it contains a partial summation result, only except in the first step, when every element in $H$ is zero. Finally, we have the same number of strides to that in the CCI case. Therefore, in the Balanced-Sparsity case, the memory I/O requirement is $(3N+G)\times N^2/G-N^2=3N^3/G$. The difference of I/O requirements between these two approaches is $N^3/G-N^2$. Since $N$ is typically much larger than $G$, CCI-Sparsity can significantly reduce the I/O complexity in this case.

\section{Constraint imposed by CCI/Balanced-Sparsity}
\label{Section:constraint}
As explained in the previous section, CCI-Sparsity is advantageous in terms of I/O efficiency for inference. However, since CCI-sparsity imposes a structural constraint on the network, it can impact overall model performance. In this section we present both theoretical and empirical analyses to assess this impact. Note that the following analysis can be similarly applied to the Balanced-Sparsity structure, leading to the same conclusion. 

\begin{figure}[htb]
\centering
\begin{subfigure}[t]{0.47\columnwidth}
\centering
   \includegraphics[width=\textwidth]{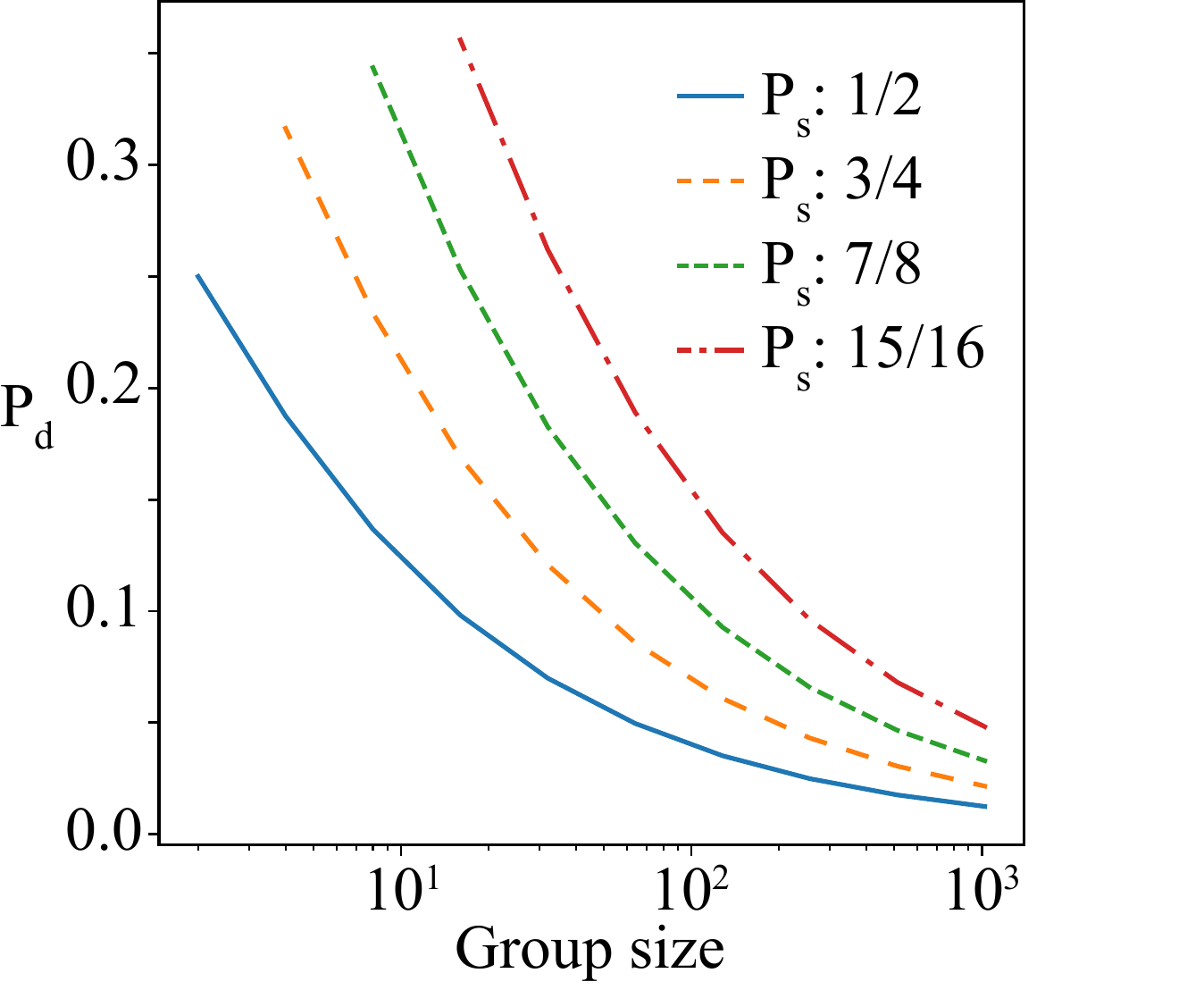}

   \caption{}
\label{fig:alloc_error}
\end{subfigure}
\hfill
\begin{subfigure}[t]{0.47\columnwidth}
\centering
   \includegraphics[width=\textwidth]{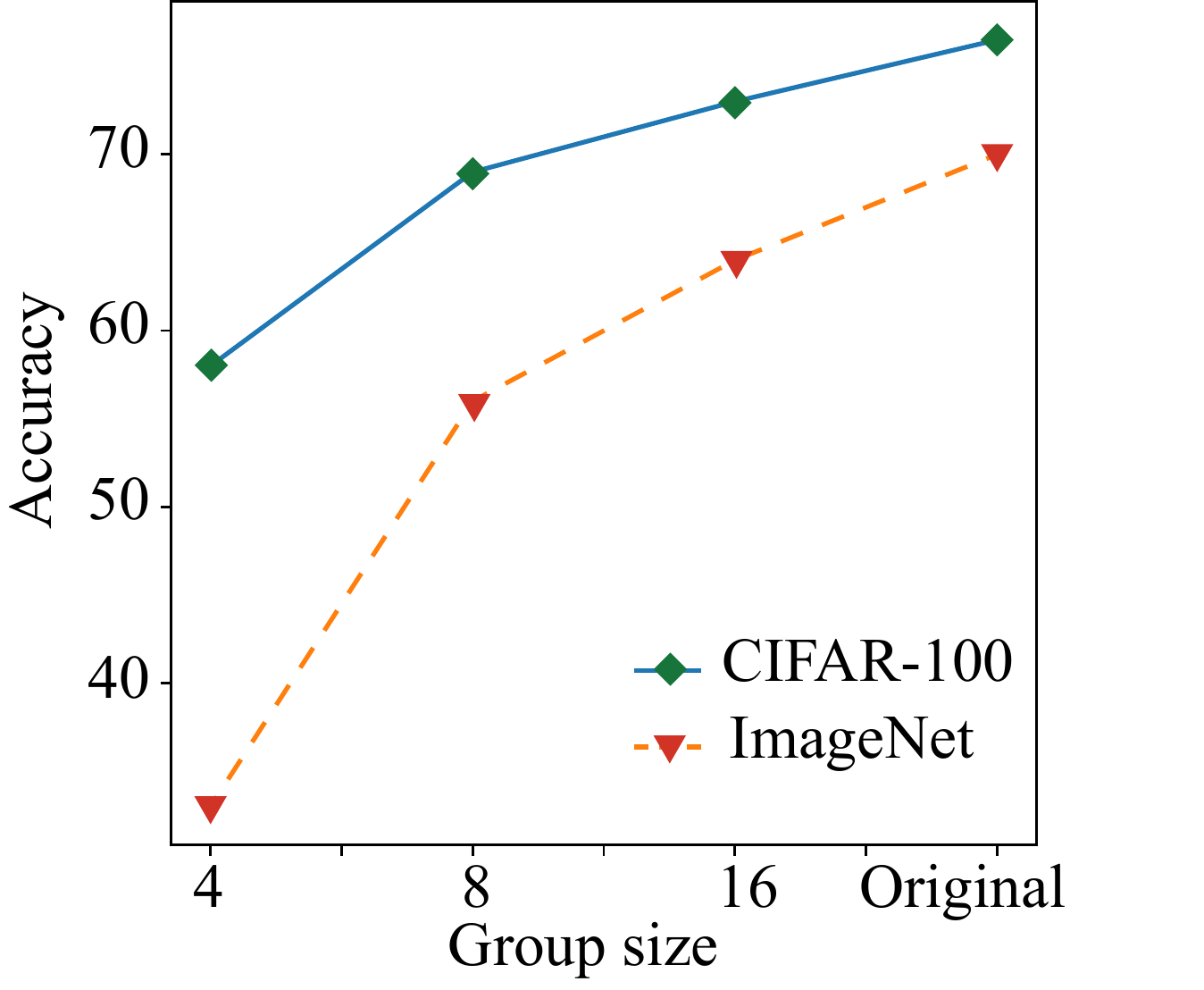}

   \caption{}
\label{fig:acc_loss}
\end{subfigure}
\caption{(a) The probabilities of a nonzero weight failing to be allocated an encoding slot under various group sizes and sparsity ratios $P_s$ of $1/2, 3/4, 7/8, \,\text{and} \,15/16$. The group sizes plotted are $\left \{ 2,4,\cdots,512,1024\right \}$. (b) The effect of group size ($G=4,8,\, \text{and}\, 16$, $s=1,2,\, \text{and}\, 4$, respectively) on the model accuracy with CCI-Sparsity imposed on a pre-trained sparse MobileNetV2 ($75\%$ sparsity ratio). ``Original" indicates the model accuracy without imposition of the CCI-Sparsity structure. We re-calibrate the batch-normalization statistics after model pruning, but no model fine-tuning is conducted. \textit{Blue solid line:} results on the CIFAR-100 dataset. \textit{Orange dashed line:} results on the ImageNet dataset.}
\label{fig:CCI_effect}
\end{figure}

Consider the case of imposing the CCI-Sparsity constraint on a pre-trained network layer of fine-grained sparsity. Again, we use a fully connected network layer for the sake of clarity. We have a weight matrix $W$ in dense form, where each row of $W$ corresponds to one output channel. Assume that each weight element in $W \in\mathbb{R}^{M\times N}$ independently has the same probability $P_s$ (sparsity ratio) of being zero. Again, we partition the $N$ weights of each row into $g=N/G$ groups, with a group size of $G$. 
Denote by $s=G\times (1-P_s)$ the number of slots that are available for the storage of the nonzero weight values in each group. For simplicity, we assume $g,G,s\in \mathbb{N}^+$. Since each weight in ${W}$ has the same probability of being nonzero, the number of nonzero weights $i$ in each weight group follows a binomial distribution $B(i,G,1-P_s)$. If $i>s$ for a weight group, then $i-s$ nonzero weights must be discarded. Denote the probability of each weight being discarded as $P_d$. Then
\begin{equation}
\label{equ:miss_prob}
  P_d=\sum_{i=1}^{G}\frac{1(i>s)(i-s)B(i,G,1-P_s)}{s}
\end{equation}
is the probability of a weight failing to be allocated to an encoding slot, where $1(\cdot)$ here is the indicator function. 
With Eqn. \ref{equ:miss_prob}, we plot the relationship among $P_d$, the group size G and the sparsity ratio $P_s$ in Fig. \ref{fig:alloc_error}. With the same sparsity ratio, $P_d$ steadily increases as the group size $G$ decreases. In addition, $P_d$ is significantly larger if we adopt a higher sparsity level with the same group size.

To gain insight into how the CCI-Sparsity group size might affect model accuracy, we apply the CCI-Sparsity structure on a pre-trained MobileNetV2 network with a sparsity ratio of $75\%$ (sparse on pointwise layers only, since most of computing is conducted in the pointwise layers). As illustrated in Fig. \ref{fig:acc_loss}, while we aim at enforcing the same sparsity ratio of $75\%$ on models, a smaller group size causes more severe performance degradation, which coincides with the rising allocation error rate when we decrease the group size. 

\section{Model training by targeted dropout} \label{Section:drop}
The result in Fig. \ref{fig:acc_loss} is obtained by post-training imposition of CCI-Sparsity on pre-trained fine-grained sparse models, \textit{i.e.} models that are not optimized with the CCI-Sparsity constraint during training. 
In this section, we present an algorithm that trains CCI-Sparse networks from scratch. The technique can also be applied to models with Balanced-Sparsity.   

There are two classes of approaches to training a sparse neural network. The first is to prune a pre-trained dense network, followed by post-training (PT) fine-tuning \cite{han2015learning,zhu2017prune,yao2019balanced}; this approach is used in Yao \etal \cite{yao2019balanced}. The other is to learn the sparse structure directly; targeted dropout (TD) \cite{gomez2018targeted} and dynamic sparse reparameterization \cite{mocanu2018scalable,mostafa2019parameter} belong to this category. Both approaches are known to yield regular sparse neural network models of satisfactory performance.

To induce fine-grained network sparsity, we select a certain number of elements out of a large group of weights, as in most network pruning methods. The weight group is typically large; for example, the group can be composed of all weights from a network layer, or even all weights from a network. For CCI-Sparsity models, the sparsity is constructed with the group constraint; that is, only limited slots are available for nonzero weights encoding inside each group. With a desirable small group size, for example, a group size of 8, this imposes a strong constraint on our models (see Fig. \ref{fig:CCI_effect}). Since the probability of any important weights being pruned away is high with small group sizes, we hypothesize that models would have higher accuracy if given more opportunities to recover pruned weights and to better adjust to the CCI-Sparsity constraint. In contrast, with the iterative PT approach, a pruned weight will not have any chance of being recovered, our hypothesis predicts that this approach would yield models of inferior performance. We test the hypothesis with experiments described in the following. 

\subsection{Improved targeted dropout}
We propose an improved version of TD for our model training. 
TD training begins with a dense network structure. During the course of training, a set of candidate network connections are selected based on a specified policy (in our case, we sort weights of low absolute magnitude for pruning). Then, the candidate connections are dropped with a specified probability similar to the dropout technique \cite{srivastava2014dropout}. 
In the original TD paper \cite{gomez2018targeted}, the authors increase the size of the dropout candidate pool during training and use a fixed $50\%$ dropout ratio. While their approach was successful on the small datasets reported, we find that this strategy does not yield models of competitive accuracy when applied to models on the ImageNet dataset \cite{ILSVRC15}. Gomez \etal \cite{gomez2018targeted} suggest that TD causes the unimportant connections in the pruning pool to shrink toward zero due to $\ell_2$ regularization. This property is important, as the weights in the pruning pool participate in the model training with a fixed probability, and if those weights are of significant magnitude, they would affect the running mean and variance statistics of the batch-normalization layers in the network, leading to inferior model performance \cite{li2018understanding}.

\begin{figure}[htb]
\centering
\begin{subfigure}[t]{0.55\columnwidth}
\centering
   \includegraphics[width=\textwidth]{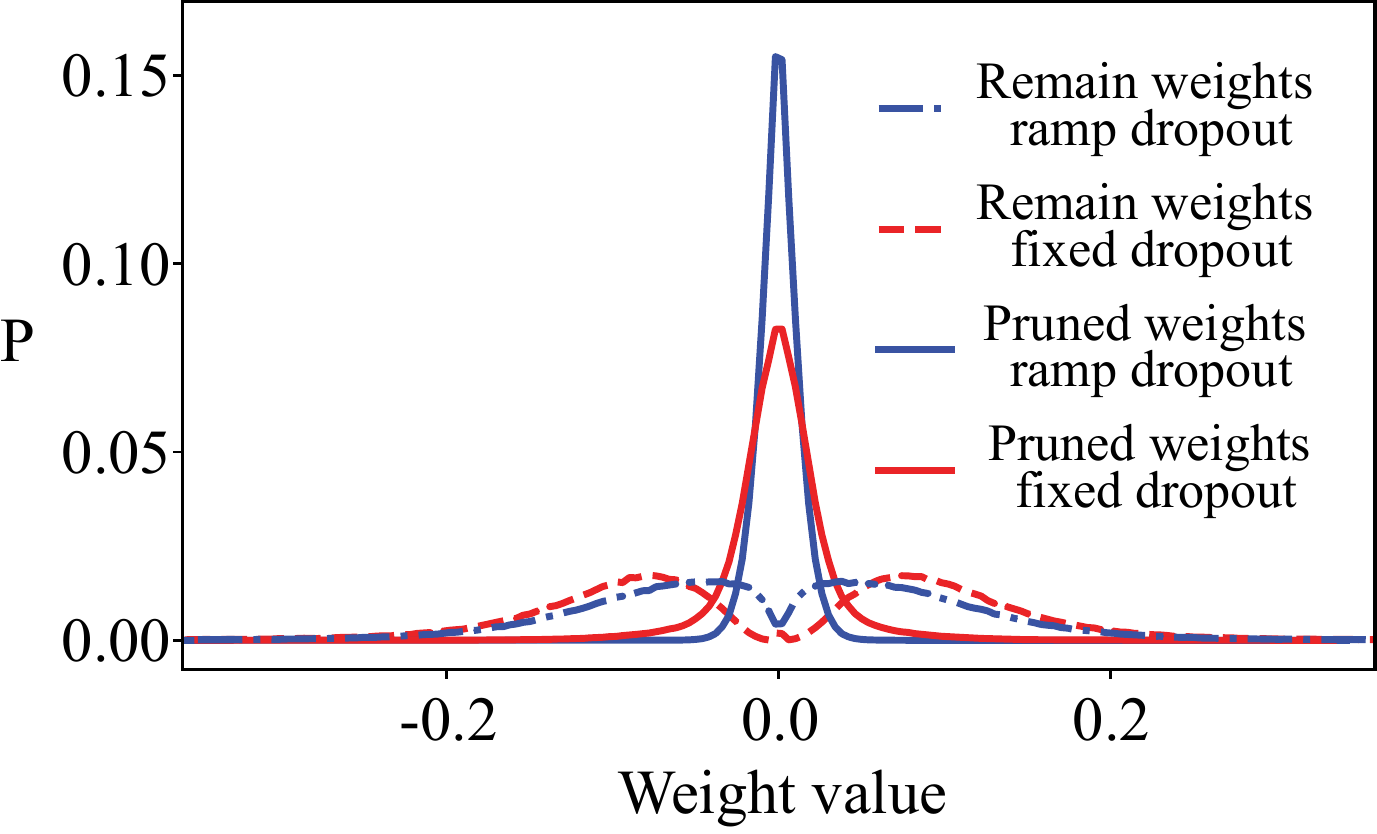}
   \caption{}
   \label{fig:weight_distribution}
\end{subfigure}
\hfill
\begin{subfigure}[t]{0.415\columnwidth}
\centering
   \includegraphics[width=\textwidth]{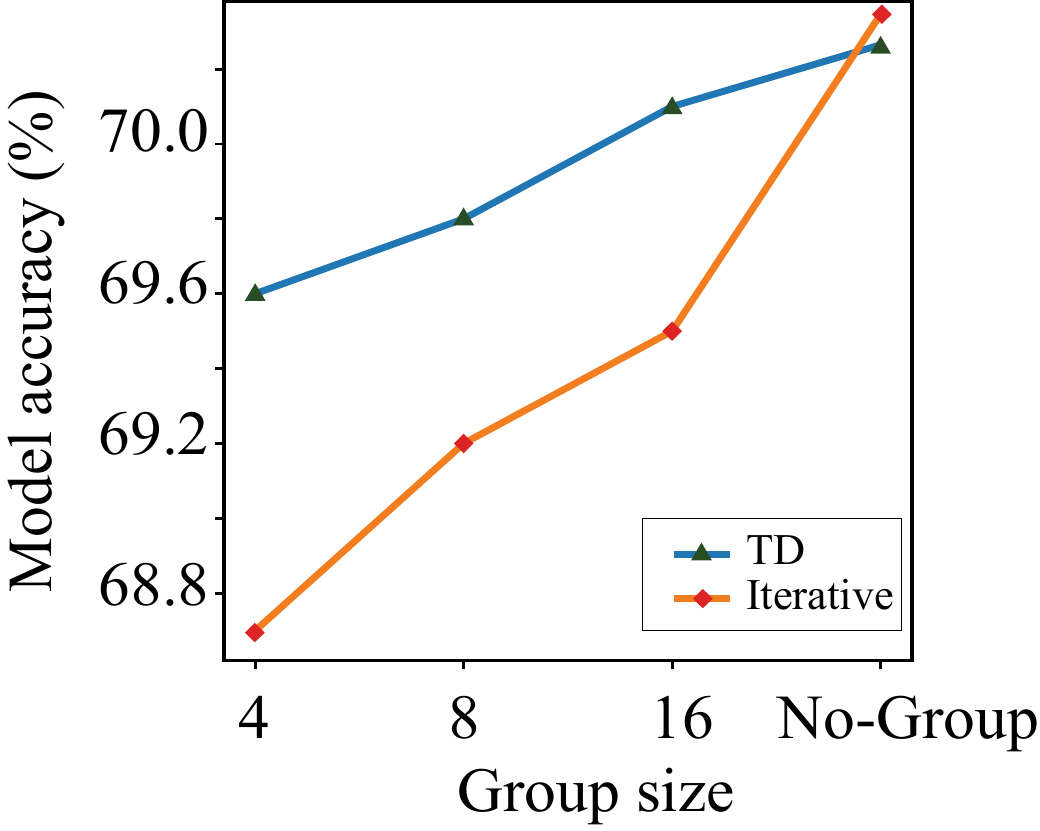}
   \caption{}
   \label{fig:groupsize}
\end{subfigure}

   \caption{(a) The weight distributions under the original (red) and proposed (blue) dropout ramping during training. Solid line: weights from the pruning pool. Dashed line: weights protected from pruning. The dropout ratio ramping causes more weights in the pruning pool to shrink toward 0. (b) The effect of the group size on model accuracy and a comparison between TD and PT pruning. "No-Group": no CCI-Sparsity imposed.} 
\label{fig:combined_2}

\end{figure}

As shown in Fig. \ref{fig:weight_distribution}, the pruned weights are of a significant magnitude in our model. Ideally, we would prefer those pruned weights to stay at the value of exact zero such that they would not affect batch normalization. A possible solution is to use a higher dropout rate for the TD-training, which allows weights in the pruning pool a higher chance to shrink toward zero. However, this also lowers the chance of recovery of a pruning candidate weight. Hence, here we increase the candidate dropout rate from the initial $50\%$ toward $100\%$ to encourage the pruning candidate weights to shrink towards zero while allowing the weights a higher chance of recovery at the early stage of the model training process. As shown in Fig. \ref{fig:weight_distribution}, the dropout rate ramping indeed causes more pruned weights to shrink toward zero. This approach leads to significantly improved model test accuracy. In a typical case with $G=8\, \text{and} \,s=2$, \textit{i.e.} a $75\%$ sparsity ratio, a MobileNetV2 model trained on the ImageNet dataset reaches a Top-1 accuracy of $69.8\%$ (ramping dropout) vs. $64.8\%$ (fixed 50\% dropout rate).

\subsection{Choices of group size for targeted dropout vs. iterative pruning}
For CCI/Balanced-Sparsity, small group sizes require lower weight index storage, as well as smaller register file sizes, leading to superior data locality or smaller area and less wiring in hardware design and a lower register file accessing energy cost \cite{yang2018dnn}. 
However, as explained in Sec. \ref{Section:CI}, small group sizes impose a strong constraint on models and thus lead to inferior performances. How do TD and PT training compare with each other in tolerating small group sizes?  

To address this question, we train MobileNetV2 models with all pointwise layer sparsity ratio set to 75\% under various group sizes, with TD and with PT. As longer training typically lead to higher model accuracy, we selected the number of training epochs such that model in the no-group setting has very similar accuracies under TD and PT to simplify the comparison. For TD, all models are trained for 120 epochs. For PT, the models are first trained for 120 epochs and iteratively pruned for additional 60 epochs. As shown in Fig. \ref{fig:groupsize}, smaller weight group sizes indeed cause more accuracy loss than a large group size (similar behavior is observed with Balanced-Sparsity, not shown). However, more importantly, even though PT yields similar accuracy to TD when no CCI-Sparsity is imposed (no-group), accuracy of models trained with TD degrade much more gracefully than those trained with PT, even if one performs 120 extra epochs of iterative pruning in the case of PT.

\section{Experimental results}
\label{Section:coarse_comp}

In this section, we investigate the characteristics of the CCI-Sparsity architecture empirically, aiming at producing networks with state-of-the-art inference efficiency at certain accuracies. Experiments are conducted on ImageNet \cite{ILSVRC15} and CIFAR-10 datasets \cite{krizhevsky2009learning}. To investigate how CCI-Sparsity affects various network architectures, we prune both heavyweight models—VGG and ResNet—and a lightweight model— MobileNetV2. 

For ImageNet experiments, we use the ILSVRC-2012 subset. The models are trained with the augmented standard training set and evaluated with the center crop of images from the validation set, as described in MobileNetV2 \cite{sandler2018mobilenetv2}. Unless explicitly specified, the models are trained with the SGD optimizer with momentum of 0.9 and an initial learning rate of 0.18, on 4 GPUs with a batch size of 64 per GPU for a total of 120 epochs with the cosine decay schedule \cite{loshchilov2016sgdr}. For standard models, the weight decay is set to 0.00004. For fair comparison, we do not tune any of the hyperparameters that are described above throughout our experiments. For TD training, we ramp up the targeted rate while keeping the candidate dropout rate at 50\% during the first half of training epochs, and then ramp up the candidate dropout rate from 50\% to 100\% during the second half. We compress convolutional layers in a network with the same group size $G$ and $s$ setting. For MobileNetV2, we only compress the pointwise convolutional layer. For VGG, we compress the fully connected layers in addition to convolutions.

For the experiments on the CIFAR-10 dataset, we perform model training as described in He \etal  \cite{he2016deep}. We train all models for a total of 400 epochs at a batch size of 64, an initial learning rate of 0.1 with a cosine decay schedule \cite{loshchilov2016sgdr}. The weight decay is set to 0.0001. We run each experiment 5 times and report the median test accuracy. In first 200 epochs, the target rate of model increase from 0 to the target value which is related to the sparsity level we want to reach and the group size we have chosen. During this process, the dropout rate remained at 50\%. In the remained epochs, the dropout rate ramping from 50\% to 100\% gradually.

\subsection{Results on the CIFAR-10 dataset}
The results of ResNet-32 and VGG-16 models on CIFAR-10 dataset are presented in Tables \ref{tab:cifar-10-ResNet} and \ref{tab:cifar-10-VGG}. Our improved TD training process yields better performance than the original TD approach \cite{gomez2018targeted}. Comparing models with CCI-Sparsity constraint against unconstrained ones at the same sparsity level, we observe a much larger performance drop than in the lightweight MobileNetV2 model as presented in Fig. \ref{fig:groupsize}. This finding supports our analysis in Sec. \ref{Section:CI} that CCI-Sparsity constraint is mild when the sparsity level is not very high. We also compare our result of VGG-16 with a structured sparse model as in \cite{li2016pruning}, where entire filters are targeted for pruning. CCI-Sparsity also demonstrates a clear advantage in this case.

\begin{table}[htb]
\resizebox{0.85\columnwidth}{!}{
\centering

\begin{tabular}[t]{@{}llll@{}}

\toprule
Model                        & \begin{tabular}[c]{@{}l@{}}Param \\ Pruned (\%)\end{tabular} & \begin{tabular}[c]{@{}l@{}}Flops \\ Pruned (\%)\end{tabular} & \begin{tabular}[c]{@{}l@{}}Top1 \\ Accuracy (\%)\end{tabular} \\ \midrule
Baseline                   & 0.00                                                            & 0.00                                                            & 92.98                                                         \\
$G=16\,/\,s=1$             & 93.51                                                        & 93.15                                                        & 90.56                                                         \\
Without Group 1    & 93.51                                                        & 93.15                                                        & 91.21                                                         \\
$G=64\,/\,s=1$             & 97.40                                                        & 95.50                                                        & 88.39                                                         \\
Without Group 2    & 97.40                                                        & 95.50                                                        & 89.42                                                         \\ \midrule
Original TD\cite{gomez2018targeted} & 94.00                                                        & 94.00                                                        & 88.80                                                         \\
Original TD & 97.00                                                        & 97.00                                                        & 88.67                                                         \\
Original TD & 98.00                                                        & 98.00                                                        & 88.70                                                         \\ \bottomrule
\end{tabular}

}
\caption{Test accuracy of the ResNet-32 model on the CIFAR-10 dataset. Top rows: results from various sparse structures. (Baseline: baseline non-sparse model; $G=16\,/\,s=1$: CCI-Sparse model with group size $G=16$, and $s=1$; without group 1: sparse model with same sparse density as $G=16\,/\,s=1$ but without the group structure; $G=64\,/\,s=1$: CCI-Sparse model with group size $G=64$, and $s=1$; without group 2: sparse model with same sparse density as $G=64\,/\,s=1$ but without the group structure.) 
Bottom rows: results from the original TD study \cite{gomez2018targeted}.}
\label{tab:cifar-10-ResNet}
\end{table}

\begin{table}[htb]
\resizebox{0.85\columnwidth}{!}{
\begin{tabular}[t]{ll}
\centering

&
\centering

\begin{tabular}[htb]{llll}
\toprule
Model                                                                                      & \begin{tabular}[c]{@{}l@{}}Params \\ Pruned (\%)\end{tabular} & \begin{tabular}[c]{@{}l@{}}Flops \\ Pruned (\%)\end{tabular} & \begin{tabular}[c]{@{}l@{}}Top1 \\ Accuracy (\%)\end{tabular} \\ \midrule
Baseline                                                                                    & 0.00                                                             & 0.00                                                            & 93.55                                                         \\

$G=16\,/\,s=1$                                                                             & 93.68                                                          & 93.21                                                         & 92.55                                                         \\
Without Group 1                                                                    & 93.68                                                          & 93.21                                                         & 92.90                                                         \\
$G=64\,/\,s=1$                                                                             & 98.36                                                          & 97.87                                                         & 90.48                                                         \\
Without Group 2                                                                    & 98.36                                                          & 97.87                                                         & 91.46                                                         \\ \midrule
$G=16\,/\,s=4$                                                                               & 74.93                                                          & 74.58                                                         & 93.43                                                        \\
\begin{tabular}[c]{@{}l@{}}VGG-16 \cite{li2016pruning})\end{tabular} & 64.00                                                            & 34.20                                                         & 93.40                                                          \\ \bottomrule
\end{tabular}
\end{tabular}
}
\caption{Test accuracy of the VGG-16 model on the CIFAR-10 dataset. Top rows: results from various sparse structures. (Baseline: baseline non-sparse model; $G=16\,/\,s=1$: CCI-Sparse model with group size $G=16$, and $s=1$; without group 1: sparse model with the same sparse density as $G=16\,/\,s=1$ but without the group structure; $G=64\,/\,s=1$: CCI-Sparse model with group size $G=64$, and $s=1$; and without group 2: sparse model with the same sparse density as $G=64\,/\,s=1$ but without the group structure.)\\
Bottom rows: comparison between CCI-Sparsity and structured sparsity \cite{li2016pruning}.}
\label{tab:cifar-10-VGG}
\end{table}

\subsection{Results on the ImageNet dataset}

\begin{figure}[t]
\centering
 \includegraphics[width=0.88\linewidth]{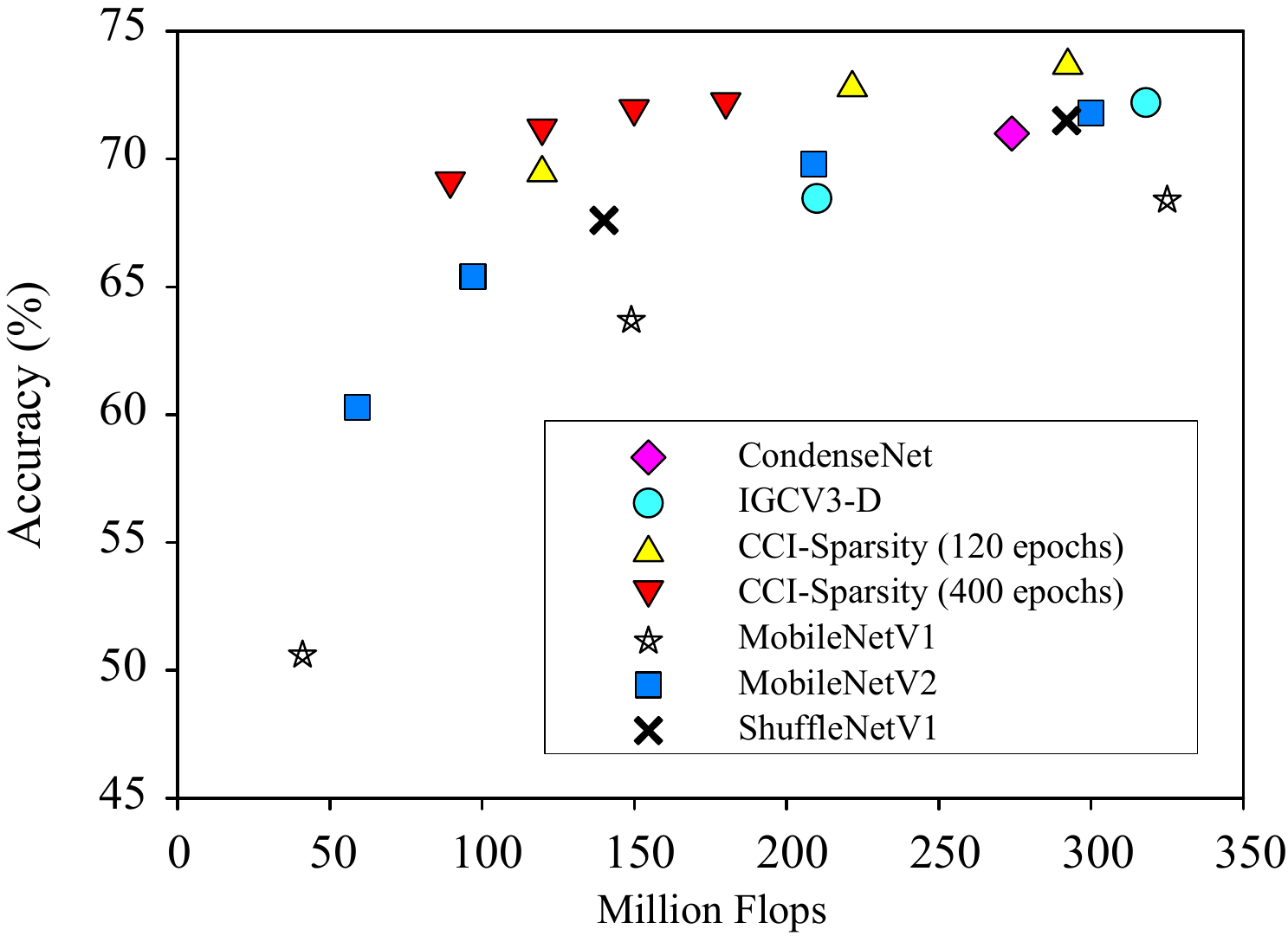}


   \caption{Efficiency-accuracy tradeoff for the ImageNet classification task compared among various lightweight or pruned network architectures. The horizontal axis corresponds to the total operation counts (in FLoPs). For MobileNetV1, MobileNetV2, and ShuffleNetV1, we also include the results from models with the width multiplied by various scales. MobileNetV2 with CCI-Sparsity: \textit{Yellow triangles:} $G=8\,/\,s=2$, with width scales of 1.0, 1.4 and 2.0, trained for 120 epochs; \textit{Red inverted triangles:} $G=8$ with $s=1,2,3 \,\text{and}\, 4$, trained for 400 epochs. 
   }
   \label{fig:net_comp}
   
\end{figure}

In this experiment, we first compare the test accuracy of CCI-Sparse models (pruned from MobileNetV2, pointwise layer only) with that of lightweight architectures from the latest literature \cite{huang2018condensenet,sun2018igcv3,ma2018shufflenet,howard2017mobilenets} in Fig. \ref{fig:net_comp}a. From these reports, the results of different model sizes are shown here. For comparison, we present results where we scale up the model size (yellow triangles), which we train for 120 epochs as in standard setting. We also show results (red inverted triangles) from models of different sparsity levels (trained for 400 epochs for better performance). Evidently, CCI-Sparse models trained with improved TD (ours) outperform those lightweight architectures by a significant margin (See Tab. \ref{tab:lightweight} in appendix for details). 
We also compare CCI-Sparsity with structured sparsity on ResNet-18 model \cite{Dong_2017_CVPR,DBLP:journals/corr/abs-1808-06866} trained on ImageNet, as shown in Tab. \ref{tab:net_comp}. Again, ours significantly outperforms structured sparsity by a significant margin. 

\begin{table}[htb]
\centering
\resizebox{0.9\columnwidth}{!}{
\begin{tabular}[b]{@{}lllll@{}}
\toprule
Method & \begin{tabular}[c]{@{}l@{}}Baseline Top1\\ Accuracy(\%)\end{tabular} & \begin{tabular}[c]{@{}l@{}}Flops \\ Pruned (\%)\end{tabular} & \begin{tabular}[c]{@{}l@{}}Top1\\ Accuracy (\%)\end{tabular} & \begin{tabular}[c]{@{}l@{}}Accuracy\\ Drop(\%)\end{tabular} \\ \midrule
Ours   & 71.12                                                            & 70.55                                                        & 70.31                                                        & 0.81                                                             \\
Dong \etal \cite{Dong_2017_CVPR}       & 69.98                                                            & 34.6                                                         & 66.33                                                        & 3.65                                                             \\
He \etal \cite{DBLP:journals/corr/abs-1808-06866}      & 70.28                                                            & 41.8                                                         & 67.10                                                        & 3.18                                                             \\ \bottomrule
\\
\\

\end{tabular}
}

   \caption{Performance comparison between two sparsity configurations on the ResNet-18 model. CCI-Sparsity (G=16\,/\,s=2) outperforms structured sparsity by a large margin.} 
   \label{tab:net_comp}
\end{table}

\section{Conclusions and future directions} \label{Section:conclusion}

This work presents the CCI-Sparsity structure and an effective algorithm to train models under such constraints. Our method retains the performance advantage of fine-grained sparsity over the coarse-grained structured pruning approach, while at same time, CCI-Sparsity avoids the inference inefficiency of fine-grained sparsity caused by irregularity in the computing dataflow. Additionally, CCI-Sparsity provides compatibility with the matrix-tiling that is often used in accelerator designs, enabling higher inference efficiency than fine-grained sparsity through data reuse and parallelism.

Through theoretical and empirical analyses, we demonstrate the trade-off between the strength of structural constraint and the computational efficiency due to the choice of group size $G$. While a small group size incurs lower computing overhead, it will have a stronger impact on the model performance than a large group size. According to our experiments, a weight group size of 16 does not typically cause significant performance loss when compared with regular sparsity cases.
Our analysis also shows that models with a higher sparsity ratio are more strongly affected by the group structure. This result suggests that one should select compact models with a proper sparsity ratio instead of very wide network models with very high sparsity ratio.
We also compare CCI-Sparsity with Balanced-Sparsity and present an important case where CCI-Sparsity requires lower I/O access than Balanced-Sparsity for model inference.

When trained with the same training algorithm, we observe no performance difference between CCI-Sparsity and Balanced-Sparsity if the same hyperparameter settings are used. Our proposed TD training strategy outperforms iterative pruning as used by Yao \etal \cite{yao2019balanced} in training models with the CCI/Balanced-Sparse models. Finally, compared with several lightweight network architectures and structural pruned models, sparse networks produced by our method outperforms the state-of-the-art in terms of best classification accuracy under a certain level of computational budget.

\paragraph{Exclusive Lasso regularization:} 
Appropriate model regularization can often lead to improved generalization performance \cite{li2016pruning,wen2016learning}. For the main part of this study, we use a simple $\ell_2$ regularizer and rely on TD to induce the intragroup sparsity structure. According to our preliminary investigation in Appendix \ref{Section:exclusive}, exclusive Lasso regularization can improve the model training and yield higher model test accuracy. Further investigation is necessary to understand the interaction between exclusive Lasso regularization and TD training. 

\paragraph{Group size:} 
As shown in Fig. \ref{fig:groupsize}, a group size of 16 can yield an accuracy close to that of the regular fine-grained sparse model for lightweight network architectures. However, with a group size of 16, the lowest sparsity ratio would be ${1}/{16}$, which might not be sufficient for compressing some of the huge networks. Moreover, according to the analysis in Section \ref{Section:CI}, CCI-Sparsity does not perform well with a very high sparsity ratio. Therefore, we believe that it is best to apply CCI-Sparsity on network models that do not contain much redundancy in the first place. How to optimally combine CCI-Sparsity with structured pruning remains an open question.

\small
\bibliography{cvpr.bib}
\bibliographystyle{ieee_fullname}
\normalsize
\clearpage
\appendix
\section{Appendix} \label{Section:supp}

\setcounter{figure}{0} 
\setcounter{table}{0} 
\setcounter{algorithm}{0}
\setcounter{equation}{0}
\renewcommand{\thefigure}{A.\arabic{figure}}
\renewcommand{\thetable}{A.\arabic{table}}
\renewcommand{\thealgorithm}{A.\arabic{algorithm}}
\renewcommand{\theequation}{A.\arabic{equation}}

\subsection{Exclusive Lasso regularization} \label{Section:exclusive}

Exclusive Lasso regularization encourages competition between components inside a group \cite{zhou2010exclusive} through applying the following regularizer:
\begin{equation}
    \ell(W)=\sum_{j=1}^{d}\left ( \sum_{k=1}^{G} {\left | W_k^j \right |}\right )^2
\end{equation}

\begin{figure}[htb]
\centering
   \includegraphics[width=0.8\columnwidth]{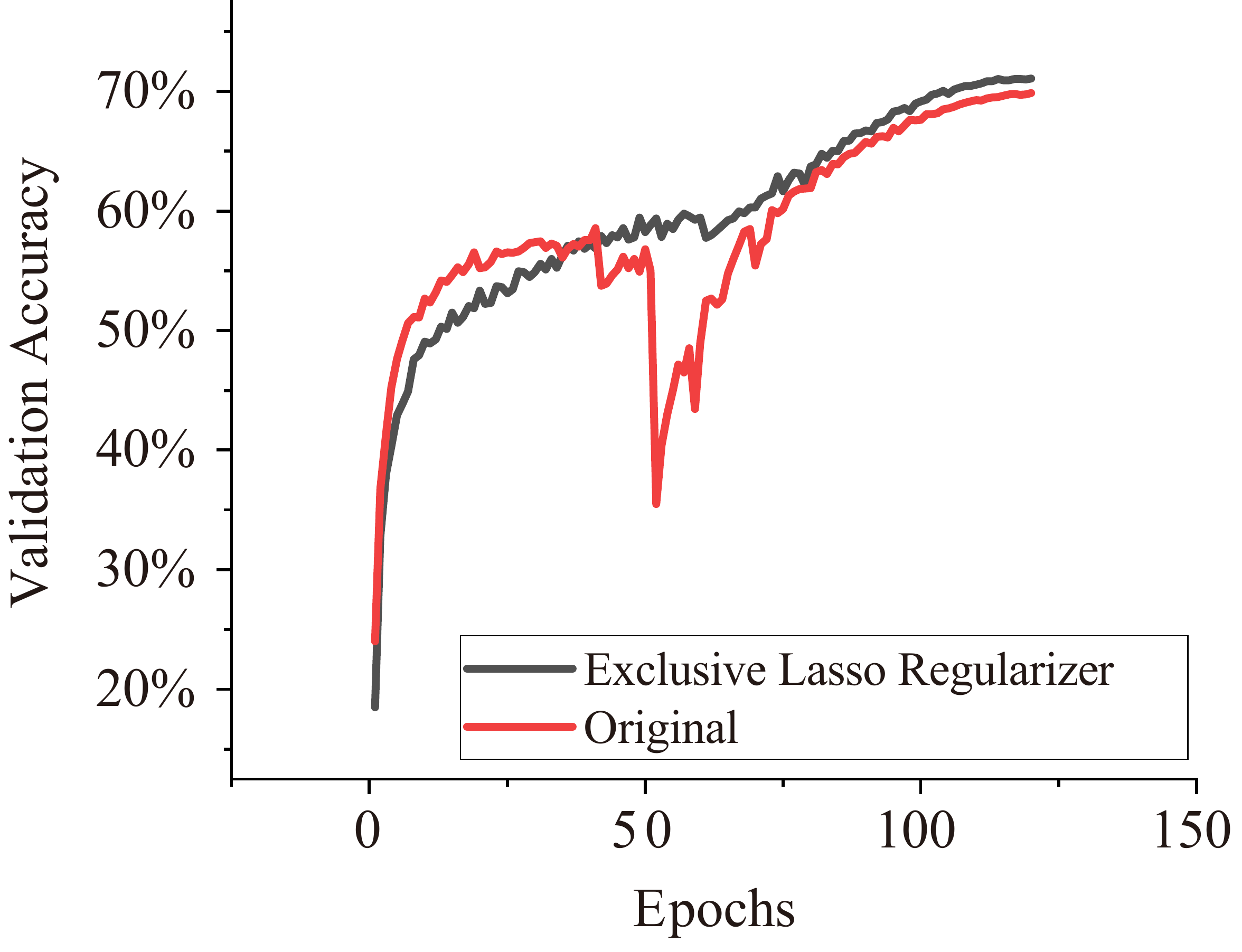}
   \caption{The exclusive Lasso regularizer improves the TD model training.}
\label{fig:Exclusive_Influence}
\end{figure}

By using the ${\ell}_{1}$ norm to combine the weights from the same group, which tends to give sparse solution, and $\ell_2$ norm to combine different groups together, which tends to minimize the regularizer loss, the exclusive Lasso regularizer essentially encourages the weights inside a group to compete for non-zero weights positions \cite{zhou2010exclusive}. Such a property is desirable for models with CCI-Sparsity. We performed experiments on combining the exclusive Lasso regularization with CCI-Sparsity. When we use the exclusive Lasso regularizer alone for CCI-Sparsity models training, we find that models generally perform rather poor. However, when we combine exclusive Lasso regularizer with the TD training, we observe a much smoother transition in the validation accuracy curve when training ramp reaches the last step, as shown in Fig. \ref{fig:Exclusive_Influence}, indicating better model convergence. We also observe that a better final model validation accuracy than a model trained with TD ( with $\ell_2$ norm regularizer) can be achieved through tuning exclusive Lasso regularization strength. As this paper is more about the testing of concept instead of achieving state of the art results, the exclusive Lasso regularizer is not included in the main text.

\newpage

\begin{table}[htb]
\centering
\resizebox{0.75\linewidth}{!}{%
\begin{tabular}{@{}llll@{}}
\toprule
\multicolumn{1}{c}{Model}                                   & \multicolumn{1}{c}{Params} & \multicolumn{1}{c}{Flops} & \multicolumn{1}{c}{\begin{tabular}[c]{@{}c@{}}Top1 \\ Accuracy\%\end{tabular}} \\ \midrule
MobileNet V1                                                & 4.2M                       & 569M                      & 70.60                                                                          \\
MobileNet V1 (0.75)                                         & 2.6M                       & 325M                      & 68.30                                                                          \\
MobileNet V1 (0.5)                                          & 1.3M                       & 149M                      & 63.70                                                                          \\ \midrule
MobileNet V2                                                & 3.4M                       & 300M                      & 72.00                                                                          \\
MobileNet V2 (0.75)                                         & 2.61M                      & 209M                      & 69.80                                                                          \\
MobileNet V2 (0.5)                                          & 1.95M                      & 97M                       & 65.40                                                                          \\ \midrule
IGCV3-D                                                     & 3.5M                       & 318M                      & 72.20                                                                          \\
IGCV3-D (0.7)                                               & 2.8M                       & 210M                      & 68.45                                                                          \\ \midrule
Condense (G=C=8)                                            & 2.9M                       & 274M                      & 71.00                                                                          \\ \midrule
ShuffleNet 1.5* (g = 3)                                     & 3.4M                        & 292M                      & 71.50                                                                          \\
ShuffleNet 1* (g = 8)                                       &                            & 140M                      & 67.60                                                                          \\
ShuffleNet 0.5* (shallow, g = 3)                            &                            & 40M                       & 57.20                                                                          \\ \midrule
CI-Sparsity on MobilenetV2 (width=1, epoch=120, G=8, S=2)   & 2.2M                       & 120M                   & 69.46                                                                          \\
CI-Sparsity on MobilenetV2 (width=1.4, epoch=120, G=8, S=2) & 3.5M                       & 222M                   & 72.78                                                                          \\
CI-Sparsity on MobilenetV2 (width=2.0, epoch=120, G=8, S=2) & 5.2M                       & 292M                   & 73.65                                                                          \\ \midrule
CI-Sparsity on MobilenetV2 (width=1, epoch=400, G=8, S=4)   & 2.61M                      & 180M                      & 72.25                                                                          \\
CI-Sparsity on MobilenetV2 (width=1, epoch=400, G=8, S=2)   & 2.19M                      & 120M                      & 71.21                                                                          \\
CI-Sparsity on MobilenetV2 (width=1, epoch=400, G=8, S=1)   & 1.97M                      & 90M                       & 69.15                                                                          \\ \bottomrule
\end{tabular}%
}
\caption{Benchmark results on lightweight architectures}
\label{tab:lightweight}
\end{table}
\end{document}